\documentclass[10pt,twocolumn,letterpaper]{article}

\usepackage[pagenumbers]{cvpr} 

\definecolor{cvprblue}{rgb}{0.21,0.49,0.74}
\definecolor{cvprred}{rgb}{0.74, 0.21, 0.21}
\usepackage[pagebackref,breaklinks,colorlinks,allcolors=cvprblue]{hyperref}
\usepackage{multicol}
\usepackage{multirow}
\usepackage{hhline}
\usepackage{booktabs}
\usepackage{caption}
\usepackage{gensymb}
\usepackage{subcaption}
\usepackage{colortbl}
\definecolor{best}{rgb}{0.96, 0.57, 0.58}
\definecolor{second}{rgb}{0.98, 0.78, 0.57}
\newcommand{\tablestyle}[2]{\setlength{\tabcolsep}{#1}\renewcommand{\arraystretch}{#2}\centering\footnotesize}

\title{MoRE: 3D Visual Geometry Reconstruction Meets Mixture-of-Experts}

\author{
Jingnan Gao$^{1,2}$\footnotemark[1]\quad
Zhe Wang$^{2}$\footnotemark[1]\quad
Xianze Fang$^{2}$\quad
Xingyu Ren$^{1}$\quad
Zhuo Chen$^{1}$\quad 
Shengqi Liu$^{1}$\quad \\
Yuhao Cheng$^{1}$\quad
Jiangjing Lyu$^{2}$\footnotemark[2] \quad
Xiaokang Yang$^{1}$\quad
Yichao Yan$^{1}$\footnotemark[3]\quad
\\
$^1$Shanghai Jiao Tong University \quad
$^2$Alibaba Group \\
}

\begin{document}
\twocolumn[{
\maketitle
\begin{center}
    \vspace{-25pt}
    \textbf{\url{https://g-1nonly.github.io/MoRE_Website/}}\\[3pt]
    \centering
    \includegraphics[width=\textwidth]{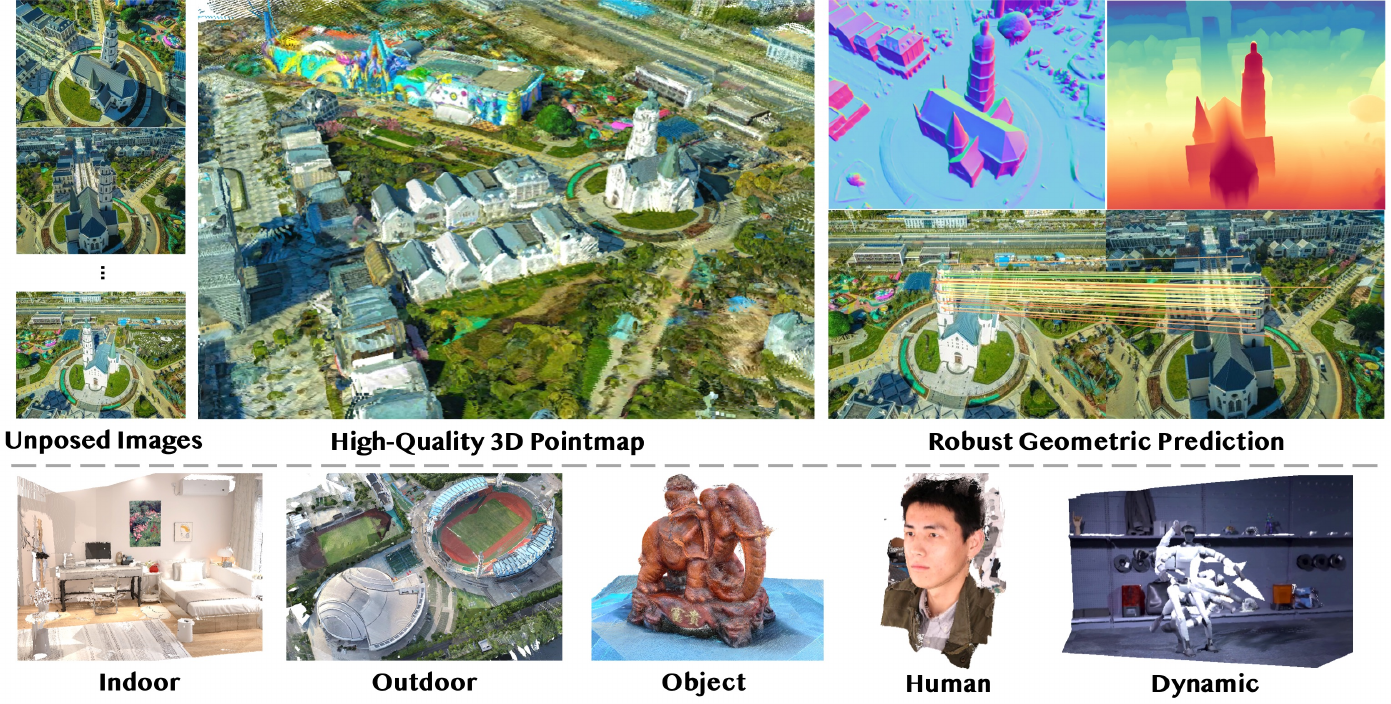}
    \vspace{-1.5em}
    \captionof{figure}{\textbf{\emph{\textcolor{cvprblue}{MoE} sparks 3D visual geometry 
\textcolor{cvprred}{R}econstruction to do 
\textcolor{cvprblue}{Mo}\textcolor{cvprred}{R}\textcolor{cvprblue}{E}.}} MoRE is a feed-forward foundation model that leverages mixture-of-experts in 3D visual geometry reconstruction. MoRE takes unposed images as input and outputs high-quality 3D pointmap, achieving robust geometric predictions for various scenarios.}
    \label{fig:teaser}
\end{center}
}
]

\let\thefootnote\relax\footnotetext{
$^*$ Equal contribution \hspace{5pt}
$^\dagger$ Project leader \hspace{5pt} 
$^\ddagger$ Corresponding author\hspace{5pt}
}

\begin{abstract}
Recent advances in language and vision have demonstrated that scaling up model capacity consistently improves performance across diverse tasks.
In 3D visual geometry reconstruction, large-scale training has likewise proven effective for learning versatile representations.
However, further scaling of 3D models is challenging due to the complexity of geometric supervision and the diversity of 3D data. 
To overcome these limitations, we propose MoRE, a dense 3D visual foundation model based on a Mixture-of-Experts (MoE) architecture that dynamically routes features to task-specific experts, allowing them to specialize in complementary data aspects and enhance both scalability and adaptability.
Aiming to improve robustness under real-world conditions, MoRE incorporates a confidence-based depth refinement module that stabilizes and refines geometric estimation.
In addition, it integrates dense semantic features with globally aligned 3D backbone representations for high-fidelity surface normal prediction.
MoRE is further optimized with tailored loss functions to ensure robust learning across diverse inputs and multiple geometric tasks.
Extensive experiments demonstrate that MoRE achieves state-of-the-art performance across multiple benchmarks and supports effective downstream applications without extra computation.
\end{abstract}

\section{Introduction}
\label{sec:intro}
Traditional methods for 3D visual-geometry reconstruction typically rely on scene-specific optimization, where models are trained for each environment or dataset.
Although such pipelines can achieve high accuracy in restricted settings, they lack the flexibility required by real-world applications that demand strong geometric priors and consistent performance across diverse scenes like AR/VR, game content creation, robotics, and autonomous driving.
To overcome these limitations, researchers draw inspiration from recent foundation models such as GPTs~\cite{gpt,gpt4}, CLIP~\cite{clip}, DINO~\cite{dino,dinov2,dinov3}, and Stable Diffusion~\cite{esser2021taming}, which demonstrate that scaling data and model capacity enables highly versatile representations across diverse tasks.
Following this paradigm, the field of 3D visual geometry reconstruction is moving beyond task-specific optimization toward scalable and generalizable architectures.
These models~\cite{dust3r,mast3r,monst3r,moge,mogev2,vggt,dens3r,fast3r,flare,vggtlong,mvdust3r,cut3r,must3r,pow3r} demonstrate the potential of feed-forward networks to enable joint 3D geometric prediction across diverse input configurations.

Notably, a key factor behind the success of these models is large-scale training, 
with evidence showing that increasing model capacity consistently yields stronger results across domains.
In the field of LLMs, scaling up Transformers has led to remarkable gains~\cite{brown2020language,hoffmann2022training,gpt4,touvron2023llama}, albeit at the cost of substantial computational resources.
A similar scaling behavior is observed in vision, where the progression from DINOv2~\cite{dinov2} to DINOv3~\cite{dinov3} highlights the benefits of enlarging model capacity for representation learning.
The same principle naturally extends to 3D visual geometry: training larger networks on larger-scale datasets can potentially improve accuracy across multiple 3D geometric prediction tasks.
However, scaling 3D models is arguably even more demanding due to the complexity of geometric supervision and heterogeneous nature of 3D data.

To address these challenges, we draw inspiration from the Mixture-of-Experts (MoE) framework, which has proven effective in scaling neural networks~\cite{fedus2022switch,lepikhin2021gshard,du2022glam,qwen2,qwen25,qwen3,deepseekmoe,DBLP:conf/acl/QiuHZWWMTL0L25}.
MoE activates only a subset of experts for each input, enabling model capacity to expand without a proportional increase in computation.
This mechanism enables experts to specialize in complementary aspects of the data, improving both scalability and adaptability.
Such specialization is particularly valuable for 3D geometry reconstruction, where scenes vary widely across indoor, outdoor, object-centric, human-centric, and dynamic environments.
By exploiting expert specialization within a unified framework, MoE provides a principled way to scale 3D geometry models effectively while maintaining computational efficiency.

In this paper, we introduce \textbf{MoRE}, a large-scale 3D visual foundation model that unifies 3D visual geometry reconstruction with the Mixture-of-Experts paradigm.
MoRE builds upon a dense visual transformer backbone and dynamically routes features to task-specialized experts, enabling the model to learn adaptive and complementary representations for different 3D scenarios.
To enhance geometric reliability, we incorporate a confidence-based depth refinement module that mitigates noise and inconsistency in real-world data.
Furthermore, we fuse dense semantic features with globally aligned 3D representations to achieve more consistent and detailed surface normal estimation.
Joint optimization across multiple 3D quantities is achieved through tailored loss functions that ensure stability and convergence during large-scale training.
Extensive experiments demonstrate that our approach achieves highly accurate 3D reconstruction and sets new state-of-the-art results across multiple benchmarks.

In summary, we make the following contributions:
\begin{itemize}
    \item We introduce \textbf{MoRE}, a dense 3D visual foundation model that leverages the \textbf{mixture-of-experts} framework in 3D geometric predictions and demonstrates high-quality performance in various 3D scenarios.
    \item We propose \textbf{dense semantic feature fusion} with \textbf{confidence-based depth refinement} to enhance the consistency and precision of 3D geometry estimation.
    \item Extensive experiments on various benchmarks showcase our \textbf{state-of-the-art} 3D geometric predictions.
\end{itemize}

\section{Related Work}
\label{sec:related}
\begin{figure*}
    \centering
    \includegraphics[width=\linewidth]{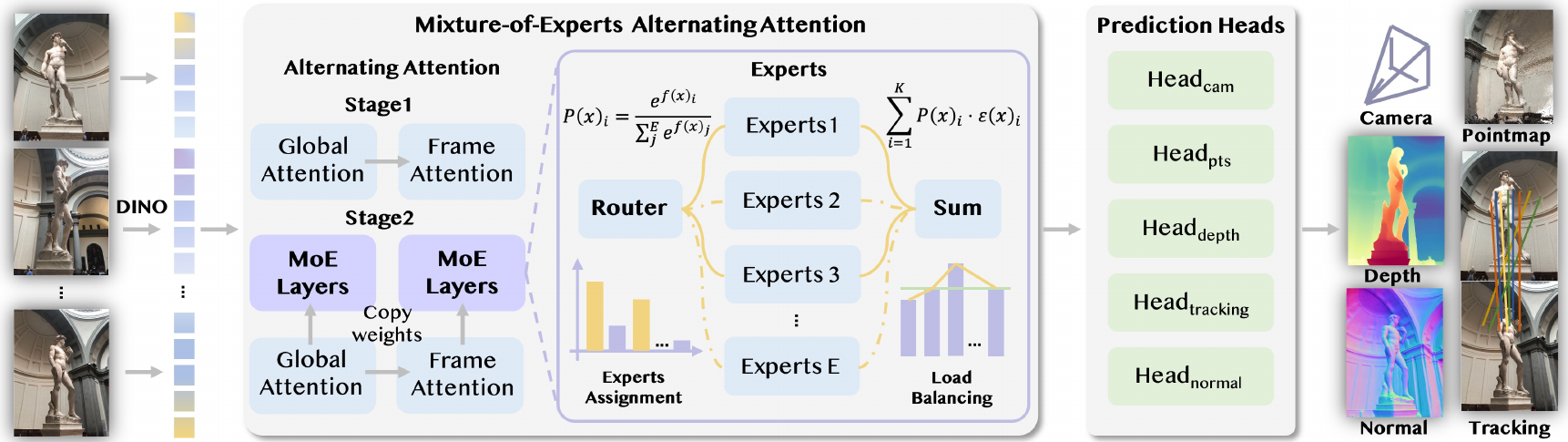}
    \caption{Overview of MoRE. We propose MoRE, a dense visual foundation model featuring a mixture-of-experts architecture and multiple task-specific heads for geometric prediction. We adopt a two-stage strategy during the model training. In Stage 1, we supervise our model with the multi-task training objectives. In Stage 2, we incorporate mixture-of-experts to further train the model for robust and accurate visual geometry reconstruction.}
    \label{fig:pipeline}
\end{figure*}
\subsection{Feed-Forward 3D Reconstruction}
With the advent of deep learning, feed-forward paradigms have emerged as a compelling alternative to traditional optimization-based pipelines for 3D reconstruction~\cite{nerf,neus,3dgs,mipnerf,Martin-BruallaR21,Barron_2023_ICCV,DBLP:conf/nips/YarivGKL21,scaffoldgs,Yu2023MipSplatting,neus2,colmap,gu2020cascade,peng2022rethinking,yao2018mvsnet,zhang2023geomvsnet,gao2025anisdf}. These methods exploit neural networks to encode strong scene priors, enabling direct regression of 3D structure from raw image inputs and improving robustness and generalization across datasets. 
Early progress was exemplified by DUSt3R~\cite{dust3r}, which produced pairwise-consistent point maps without calibration. However, its reliance on pairwise predictions necessitated a global alignment stage for larger scenes. Subsequent extensions introduced mechanisms to alleviate these limitations. MASt3R~\cite{mast3r} incorporated confidence-weighted objectives to approximate metric scale, and Fast3R~\cite{fast3r} scaled inference to thousands of views, thereby obviating alignment entirely. Other approaches aimed to reformulate the task, such as FLARE~\cite{flare}, which decouples pose estimation from geometry prediction. More recently, large-scale transformer models~\cite{vggt,moge,dens3r,mvdust3r,cut3r,must3r,pow3r} like VGGT have advanced the state of the art by jointly predicting intrinsic and extrinsic parameters, dense depth, point maps, and feature correspondences. Subsequent approaches~\cite{vggtlong,pi3} further enhanced stability and accuracy through more sophisticated strategies. Despite these advances, existing feed-forward frameworks remain limited in scalability and generalization. Our approach tackles these issues by incorporating mixture-of-experts architecture, enabling more accurate and high-fidelity reconstruction.

\subsection{Mixture-of-Experts Framework}
Recent large language models (LLMs) increasingly adopt the mixture-of-experts (MoE) framework to address the high computational cost of densely activated Transformers. As demonstrated by several early approaches~\cite{DBLP:journals/corr/abs-2305-14705,DBLP:conf/icml/XueZFNZZ024,DBLP:journals/corr/abs-2401-04088}, MoE achieves substantial efficiency gains while maintaining competitive performance by activating only a small subset of parameters during inference. Beyond efficiency, MoE~\cite{DBLP:conf/cvpr/WangBDBPLAMSSW23,3dmoe} has also been utilized for modality specialization, where different modalities are routed to dedicated experts. Several works~\cite{DBLP:conf/emnlp/0002QDRTH024,DBLP:journals/corr/abs-2401-15947,DBLP:journals/pami/LiJHWZLMZ25} have also explored converting pretrained dense LLMs into MoE models, combining the strengths of existing LLMs with the scalability and efficiency advantages of MoE. To further improve the performance across diverse downstream tasks, recent approaches~\cite{qwen2,deepseekmoe,qwen25,qwen3} have leveraged fine-grained expert segmentation and shared experts routing. 
Inspired by the effectiveness of MoE in improving prediction across diverse data domains, we propose a 3D foundation model built upon the MoE framework. Since 3D geometric reconstruction encompasses highly diverse data distributions, our approach leverages MoE to adaptively model such diversity and achieve robust geometric prediction.
\section{Method}
\label{sec:method}

This work aims to build an end-to-end framework for predicting various 3D geometric quantities from unconstrained images input. Our approach consists of three main components. First, we employ a dense visual transformer backbone to extract features that adapt to unconstrained input requirements (Sec.~\ref{sec:backbone}). 
Second, we introduce a mixture-of-experts (MoE) mechanism into the geometric prediction pipeline, allowing the model to effectively enhance both accuracy and scalability and dynamically allocate its capacity across different scenarios (Sec.~\ref{sec:moe}).
Finally, we develop specialized training strategies that stabilize optimization and further improve the generalization of the learned representations to various tasks (Sec.~\ref{sec:training}). 
With these components, our proposed model \textbf{MoRE} can flexibly adapt to varying input scenarios while achieving strong performance across a wide range of 3D geometric vision problems.
\subsection{Model Architecture}
\label{sec:backbone}
Inspired by recent progress in 3D vision~\cite{vggt,dens3r,pi3,lvsm}, our goal is to develop a foundation model that can predict a variety of geometric quantities across diverse scenes and tasks. To achieve this, we employ a dense visual transformer backbone trained on large-scale 3D annotated datasets.
Given a sequence of $N$ RGB images $(I_i)^N_{i=1} \in \mathcal{R}^{3\times H\times W}$, MoRE's dense visual transformer is a function $f$ that maps the input to a corresponding set of 3D quantities per frame:
\begin{equation}
    (C_i, P_i, D_i, T_i, N_i, )_{i=1}^N = f((I_i)^N_{i=1}),
\end{equation}
where $C_i \in \mathcal{R}^9 $ is the camera parameters including both intrinsics and extrinsics, $P_i \in \mathcal{R}^{3\times H \times W}$ is the pointmap, $D_i \in \mathcal{R}^{H \times W}$ is the depth map, $T_i \in \mathcal{R}^{C\times H \times W}$ is a grid of $C$-dimensional features for point tracking, and $N_i \in \mathcal{R}^{3 \times H \times W}$ is the normal map.
Apart from the pointmap head, depth head, camera head and tracking head implemented by VGGT~\cite{vggt}, we additionally implement a normal prediction head to facilitate normal map estimation. We further design a confidence-based depth refinement for more accurate monocular depth estimation.

\begin{figure}
    \centering
    \includegraphics[width=\linewidth]{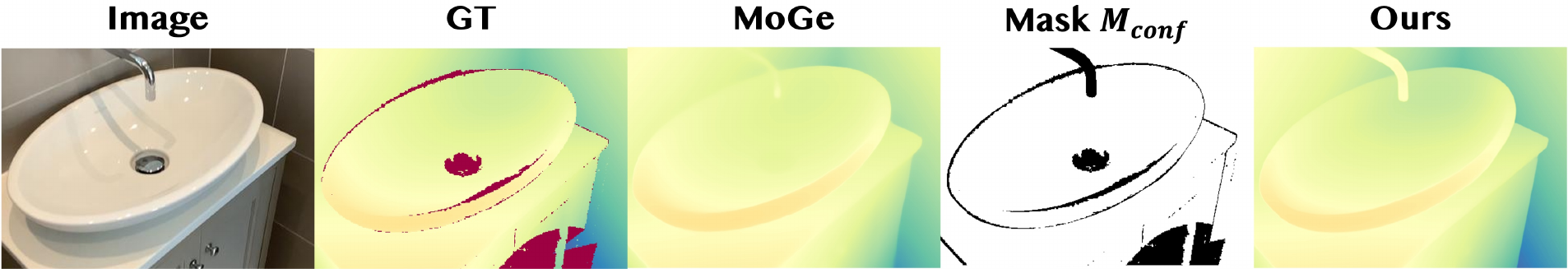}
    \caption{Real-world depth comparison. We present the ground-truth depth, prediction from MoGe, the confidence mask and our prediction after training with confidence-based depth refinement.}
    \label{fig:refd}
\end{figure}

\noindent\textbf{Confidence-based Depth Refinement.}
Real-world depth training data often contain noise and missing measurements, which can cause models to overfit unreliable depth ground truth and harm estimation accuracy.
However, a state-of-the-art monocular model~\cite{mogev2} with refined training data still produces reasonably accurate results as shown in Fig.~\ref{fig:refd}.
To further leverage this insight, we design a confidence-based depth refinement to filter depth supervision. Specifically, we utilize MoGev2~\cite{mogev2} to compute the confidence masks for each depth sample:
\begin{equation}
M_{\text{conf}} = \left\{ \frac{\left| D_{\text{moge}} - D_{\text{gt}} \right|}{\max(D_{\text{gt}}, \,\alpha)} < \tau \right\},
\end{equation}
where $D_{\text{gt}}$ is the ground-truth depth maps and $D_{\text{moge}}$ is the aligned depth map predicted by MoGev2. 
We set $\alpha=0.5$ to prevent instability from small depth values, and use $\tau=0.1$ as the threshold. 
The resulting mask is then used to filter out low-confidence, noisy, or incomplete measurements from the ground-truth depth. We incorporate this into the training by adding a prior-guided depth term to the overall depth loss $\mathcal{L}_{\text{depth}}^{\text{vggt}}$ from VGGT~\cite{vggt}:
\begin{equation}
\begin{aligned}
    \mathcal{L}_{\text{depth}}^p &= \mathcal{L}_{\text{grad}}(\hat{D}^{M_{\text{conf}}}, D_{\text{moge}}^{M_{\text{conf}}}), \\
\mathcal{L}_{\text{depth}} &= \mathcal{L}_{\text{depth}}^{\text{vggt}} + \mathcal{L}_{\text{depth}}^p
\end{aligned}
\end{equation}
where $\hat{D}^{M_{\text{conf}}}$ denotes the predicted depth maps masked by the confidence prior, and $D_{\text{moge}}^{M_{\text{conf}}}$ corresponds to the filtered depth prediction from MoGe. $\mathcal{L}_{\text{grad}}$ is the gradient-based loss as implemented in VGGT.
By restricting supervision only to high-confidence regions, our model can avoid overfitting to corrupted data, thereby achieving more accurate and stable depth estimation.

\noindent\textbf{Dense Semantic Feature Fusion.}
While monocular or binocular models~\cite{moge,mogev2,dust3r,dens3r,mast3r} can produce sharp and detailed geometry for a single view, multiview models tend to favor smoother predictions to preserve 3D consistency, which leads to the loss of fine geometric details. To address this, we fuse the globally aligned 3D feature $f_{3d}$ from the backbone with dense semantic features $f_{s}$ extracted from each input image using DINOv2~\cite{dinov2}, providing additional local geometry cues that help produce sharper and more accurate predictions.
These two features are concatenated along the feature dimension and passed through the DPT heads to regress the final depth and surface normals:
\begin{equation}
    f_{n} = f_{3d} \oplus f_{s},
\end{equation}
where $f_{n}$ is the input feature of the normal prediction head.
We empirically validated this claim in the \textit{ablation study}.

\subsection{Mixture-of-Experts Design}
\label{sec:moe}
Leveraging large vision models to predict multiple 3D geometric quantities has proven to be effective for several downstream tasks.
However, a single decoder feature may be insufficient to capture the varying domain characteristics of different 3D scenarios.
To address this limitation, we draw inspiration from the Mixture-of-Experts (MoE) framework, which has shown strong scalability and efficiency in large language models~\cite{qwen2,qwen25,qwen3,deepseekmoe,DBLP:conf/acl/QiuHZWWMTL0L25}.
In the MoE design, multiple experts act as independent sub-networks trained to capture distinct aspects of the data.
Building on this idea, we propose an MoE framework for 3D geometric vision as demonstrated in Fig.~\ref{fig:pipeline}.
This framework dynamically routes features to task-specific experts, allowing the backbone to learn specialized representations and achieve substantial improvements across downstream tasks.
Subsequently, we formulate the MoE forward pass and training objectives to fully leverage the MoE framework in 3D visual geometry reconstruction.

\noindent\textbf{MoE Forward-Pass.} 
Within the MoE framework, an MoE layer serves as a modular component that enables conditional computation and expert specialization.
A typical MoE layer comprises multiple feed-forward networks (FFNs), each serving as an expert.
For initialization, we replicate the FFNs from the alternating attention structure (global and frame attention) to construct an ensemble of experts ${\varepsilon_i}$.
A router is then employed to predict the assignment probability of each token to the corresponding experts.
In our framework, the router is implemented as a linear layer, and the routing process can be expressed as:
\begin{equation}
    \mathcal{P}(x)_i = \frac{e^{f(x)_i}}{\sum^E_j e^{f(x)_j}},
\end{equation}
where $E$ represents the number of experts and $f(x)=W \cdot x$ is the weight logits produced by the router, and the $W$ denotes the lightweight training parameters.
Each token is then processed by the top-$K$ experts with the highest assignment probabilities, and the final representation is obtained as the weighted sum according to these probabilities:
\begin{equation}
    \mathrm{MoE}(x) = \sum_{i=1}^{K} \mathcal{P}(x)_i \cdot \varepsilon(x)_i.
\end{equation}

\noindent\textbf{MoE Training Objectives.} 
Since integrating multiple experts may lead to uneven expert utilization, it is necessary to apply load-balancing constraints to the MoE layer to regularize training.
We therefore incorporate the differentiable load-balancing loss~\cite{DBLP:journals/jmlr/FedusZS22,DBLP:journals/corr/abs-2401-15947} in each MoE layer to encourage all experts to process tokens in a balanced manner:
\begin{equation}
    \mathcal{L}_{\text{moe}} = E \cdot \sum_{i=1}^{E} \mathcal{F}_i \cdot \mathcal{G}_i,
\end{equation}
where $\mathcal{F}_i$ denotes the fraction of tokens processed by each expert $\varepsilon_i$,  and $\mathcal{G}_i$ represents the average routing probability of $\varepsilon_i$, which can be formulated as:
\begin{equation}
    \begin{aligned}
        \mathcal{F} = \frac{1}{K} \sum_{i=1}^{E} &\mathbf{1} \{\mathrm{argmax}\mathcal{P}(x)=i\} , \\
         \mathcal{G}  = &\frac{1}{K} \sum_{i=1}^{K}\mathcal{P}(x)_i.
    \end{aligned}
\end{equation}

\subsection{Multi-task Training Objectives}
\label{sec:training}
Building on the training objectives of VGGT~\cite{vggt}, we initially employ the pointmap loss $\mathcal L_{\text{points}}$, the camera loss $\mathcal L_{\text{cam}}$,  and the tracking loss $\mathcal L_{\text{track}}$. We then introduce the depth loss $\mathcal L_{\text{depth}}$ after our confidence-based depth refinement and additional loss terms to further improve the accuracy and generalization of the model.

\noindent\underline{(1) Local Point Loss $\mathcal{L}_{\text {pts\_local}}$.}
Monocular geometry estimation often suffers from focal-distance ambiguity. To address this, we employ an additional local point loss to improve monocular depth estimation following~\cite{pi3,moge}. 
Given each image $\mathbf{I}_i$, a local point map $\hat{\mathbf{P}}_i$ is formed based on the predicted depth and focal parameters following VGGT~\cite{vggt}.
During training, this predicted point map can be aligned with the ground truth point map by solving for a single optimal scale factor $\hat{s}$ that minimizes the depth-weighted $\mathcal{L}_1$ distance over the entire image sequence. The optimization problem is formulated as:
\begin{equation}
\hat{s} = \underset{s}{\arg\min} \sum_{i=1}^N \sum_{j=1}^{H \times W} \frac{1}{z_{i,j}} \| s \hat{\mathbf{p}}_{i,j} - \mathbf{p}_{i,j} \|_1,
\end{equation}
where $\hat{\mathbf{p}}_{i,j} \in \mathbb{R}^3$ denotes the predicted 3D point at index $j$ of the point map $\hat{\mathbf{P}}_i$, and $\mathbf{p}_{i,j}$ is the corresponding ground-truth in $\mathbf{P}_i$. The term $z_{i,j}$ corresponds to the ground-truth depth, which is the z-component of $\mathbf{x}_{i,j}$. 
Finally, the local point cloud reconstruction loss $\mathcal{L}_{\text{points}}$ is defined based on the optimal scale factor$\hat{s}$:
\begin{equation}
\mathcal{L}_{\text{pts\_local}} = \sum_{i=1}^N \sum_{j=1}^{H \times W} \frac{1}{z_{i,j}} \| \hat{s} \hat{\mathbf{x}}_{i,j} - \mathbf{x}_{i,j} \|_1.
\end{equation}

\noindent\underline{(2) Point Normal Loss $\mathcal{L}_{\text {pts\_n}}$.}
To encourage the reconstruction of locally smooth surfaces, we employ the normal loss proposed by MoGe~\cite{moge}. For each point $\hat{\mathbf{X}}_i$ in the predicted point map, its normal vector $\hat{\mathbf{n}}_{i,j}$ is computed from the cross product of the vectors to its neighboring points on the image grid. The predicted normals are then supervised by minimizing the angular difference with their corresponding ground-truth normals $\mathbf{n}_{i,j}$:
\begin{equation}
\mathcal{L}_{\text{pts\_n}} = \sum_{i=1}^N \sum_{j=1}^{H \times W} \arccos\left(\hat{\mathbf{n}}_{i,j} \cdot \mathbf{n}_{i,j}\right).
\end{equation}

\noindent\underline{(3) Predicted Normal Loss $\mathcal{L}_{\text {n}}$.}
In addition to the pointmap normal loss, we introduce a predicted normal loss to supervise the normal head, which is responsible for predicting view-space normals:
\begin{equation}
    \mathcal{L}_{\text {n}}=\mathcal{L}_1(N,\bar{N}),
\end{equation}
where $N$ is the ground-truth normal and $\bar N$ is the view-space normal produced by the normal prediction head.

\begin{figure*}
    \centering
    \includegraphics[width=\linewidth]{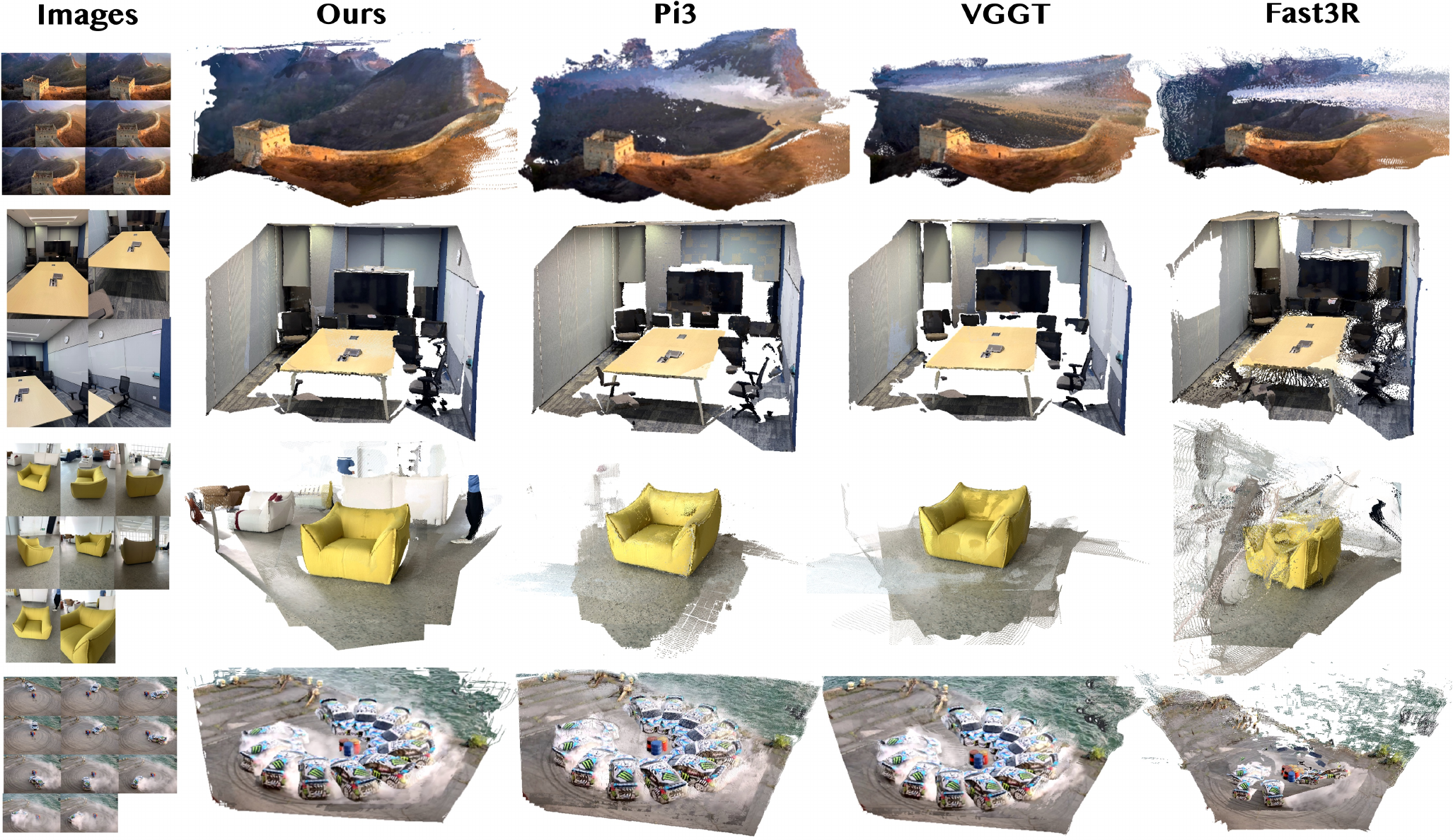}
    \caption{Qualitative comparison of multi-view 3D reconstruction. Our method demonstrates superior accuracy and robustness across diverse scenarios compared to previous feed-forward approaches.}
    \label{fig:pm}
\end{figure*}

\subsection{Model Training}
We train our model end-to-end using the following multi-task training objective:
\begin{equation}
\begin{aligned}
\mathcal L &= \mathcal L_{\text{pts}} + \mathcal L_{\text{cam}}  + \mathcal L_{\text{depth}} 
+ \lambda_{\text{track}}\mathcal L_{\text{track}}   \\
&+ \lambda_{\text{moe}}\mathcal L_{\text{moe}} + \lambda_{\text{pts\_n}}\mathcal L_{\text{pts\_local}} + \lambda_{\text{pts\_n}}\mathcal L_{\text{pts\_n}} 
+ \lambda_{\text{n}}\mathcal L_{\text{n}} 
\end{aligned}
\end{equation}
Our model is trained on a diverse dataset with varying levels of quality. 
Consequently, inaccurate annotations can occasionally cause unstable loss spikes during the training process.
To mitigate this issue, we implement an adaptive loss strategy to \textbf{stabilize training}. 
Specifically, we maintain a sliding window of recent loss values and compute their mean $\mu$ and standard deviation $\sigma$.
A dynamic threshold $T_L$ is then defined following the k-sigma rule:
\begin{equation}
    T_L =  \mu_L + k \sigma_L,
\end{equation}
where we set $k = 3$ by default in our experiments.
If the current loss exceeds this threshold $L_{cur} > T_L$, it is considered an outlier and clipped to the threshold. 
This ensures that the training is guided by the typical distribution of losses rather than being dominated by rare extreme values.
By continuously updating the loss history and applying this strategy, we effectively mitigate optimization instability while preserving the overall learning dynamics.

As for implementation details, we train our model based on the pretrained VGGT~\cite{vggt} checkpoint and 
set the hyperparameter as follows: $\lambda_{\text{moe}}=0.01, \lambda_{\text{pts\_local}}=0.5, \lambda_{\text{pts\_n}}=1.0, \lambda_{\text{n}}=1.0$.
We adopt the same training dataset as VGGT~\cite{vggt} and extend it with an internal dataset that spans indoor, outdoor, object-centric, human-centric, and dynamic scenes.
\section{Experiments}
\label{sec:exp}
We compare our method to state-of-the-art approaches on a variety of 3D geometric prediction tasks to demonstrate its effectiveness and robustness.

\subsection{Pointmap Estimation}
For pointmap evaluation, we evaluate the model on the object-centric DTU~\cite{dtu} and scene-level ETH3D~\cite{eth3d} datasets, sampling keyframes every 5 images. We further follow the evaluation protocols from~\cite{cut3r,pi3} and assess the quality of reconstructed multi-view point maps on the 7-Scenes~\cite{7scenes} and NRGBD~\cite{NRGBD} datasets. Keyframes are selected with a stride of 200 for 7-Scenes and 500 for NRGBD. 
The predicted point maps are first aligned to the ground truth using the Umeyama algorithm for coarse Sim(3) alignment, and then refined using Iterative Closest Point (ICP). We report Accuracy (Acc.), Completion (Comp.), and Normal Consistency (N.C.) in Tables~\ref{tab:mv_recon_1}.  
We also present qualitative comparisons for pointmap estimation in Fig.~\ref{fig:pm}.
The results demonstrate the effectiveness of our method across various 3D reconstruction scenarios and show consistently high performance in both sparse-view and dense-view settings. 
It is noteworthy that Pi3~\cite{pi3} often produces “checkerboard” artifacts due to insufficient learning in its transformer architecture. VGGT~\cite{vggt} and Fast3R~\cite{fast3r} tend to yield less accurate reconstructions and struggle to generalize across scenarios.
In contrast, our method reconstructs geometry that is both accurate and spatially consistent across diverse conditions.

\begin{table*}[t]
    \centering
    \tablestyle{2pt}{1.05}
    \resizebox{1.0\linewidth}{!}{
    \begin{tabular}{lccccccccccccccccccccccccccc}
        \toprule[0.17em]
        {\multirow{4}{*}{\textbf{Method}}} &
        \multicolumn{6}{c}{\textbf{DTU~\cite{dtu}}} &
        \multicolumn{6}{c}{\textbf{ETH3D~\cite{eth3d}}}  &
        \multicolumn{6}{c}{\textbf{7-Scenes~\cite{7scenes}}} &
        \multicolumn{6}{c}{\textbf{NRGBD~\cite{NRGBD}}}\\
        \cmidrule(r){2-7} \cmidrule(r){8-13} \cmidrule(r){14-19} \cmidrule(r){20-25}
        &
        \multicolumn{2}{c}{Acc. $\downarrow$}  &
        \multicolumn{2}{c}{Comp. $\downarrow$} &
        \multicolumn{2}{c}{N.C. $\uparrow$}     & 
        \multicolumn{2}{c}{Acc. $\downarrow$}  &
        \multicolumn{2}{c}{Comp. $\downarrow$} &
        \multicolumn{2}{c}{N.C. $\uparrow$}     & 
        \multicolumn{2}{c}{Acc. $\downarrow$}  &
        \multicolumn{2}{c}{Comp. $\downarrow$} &
        \multicolumn{2}{c}{N.C. $\uparrow$}     & 
        \multicolumn{2}{c}{Acc. $\downarrow$}  &
        \multicolumn{2}{c}{Comp. $\downarrow$} &
        \multicolumn{2}{c}{N.C. $\uparrow$}     &\\
        \cmidrule(r){2-3} \cmidrule(r){4-5} \cmidrule(r){6-7}
        \cmidrule(r){8-9} \cmidrule(r){10-11} \cmidrule(r){12-13}
        \cmidrule(r){14-15} \cmidrule(r){16-17} \cmidrule(r){18-19}
        \cmidrule(r){20-21} \cmidrule(r){22-23} \cmidrule(r){24-25}
        &
        Mean & Med. &
        Mean & Med. &
        Mean & Med. &
        Mean & Med. &
        Mean & Med. &
        Mean & Med. &
        Mean & Med. &
        Mean & Med. &
        Mean & Med. &
        Mean & Med. &
        Mean & Med. &
        Mean & Med. & \\
        \midrule[0.08em]
        Fast3R~\cite{fast3r} & 3.340 & 1.919 & 2.929 & 1.125 & 0.671 & 0.755 & 0.832 & 0.691 & 0.978 & 0.683 & 0.667 & 0.766 & 
        0.038 & 0.015 & 0.056 & 0.018 & 0.645 & 0.725 & 0.072 & 0.030 & 0.050 & 0.016 & 0.790 & 0.934 \\
    CUT3R~\cite{cut3r} & 4.742 & 2.600 & 3.400 & 1.316 & 0.679 & 0.764 & 0.617 & 0.525 & 0.747 & 0.579 & 0.754 & 0.848 &
            0.022 & 0.010 & 0.027 & 0.009 & 0.668 & 0.762 & 0.086 & 0.037 & 0.048 & 0.017 & 0.800 & 0.953 \\
    FLARE~\cite{flare} & 2.541 & 1.468 & 3.174 & 1.420 & \colorbox{second}{0.684} & \colorbox{second}{0.774} & 0.464 & 0.338 & 0.664 & 0.395 & 0.744 & 0.864 &
            0.018 & 0.007 & 0.027 & 0.014 & 0.681 & 0.781 & 0.023 & 0.011 & 0.018 & 0.008 & 0.882 & 0.986 \\
    VGGT~\cite{vggt} & 1.338 & 0.779 & 1.896 & 0.992 & 0.676 & 0.766 & 0.280 & 0.185 & 0.305 & 0.182 & 0.853 & 0.950 &
            0.022 & 0.008 & 0.026 & 0.013 & 0.663 & 0.757 & 0.017 & 0.010 & 0.015 & 0.005 & 0.893 & \colorbox{second}{0.988}\\
    Pi3~\cite{pi3} & \colorbox{second}{1.198} & \colorbox{second}{0.646} & \colorbox{second}{1.849} & \colorbox{best}{0.607} & 0.678 & 0.768 & \colorbox{best}{0.194} & \colorbox{best}{0.131} & \colorbox{best}{0.210} & \colorbox{best}{0.128} & \colorbox{best}{0.883} & \colorbox{second}{0.969} &
            \colorbox{second}{0.015} & \colorbox{second}{0.007} & \colorbox{best}{0.022} & \colorbox{second}{0.011} & \colorbox{second}{0.687} & \colorbox{second}{0.790} & \colorbox{second}{0.015} & \colorbox{second}{0.008} & \colorbox{best}{0.013} & \colorbox{second}{0.005} & \colorbox{second}{0.898} & 0.987 \\
         Ours & \colorbox{best}{1.011} & \colorbox{best}{0.584} & \colorbox{best}{1.491} & \colorbox{second}{0.630} & \colorbox{best}{0.695} & \colorbox{best}{0.782} & \colorbox{second}{0.234} & \colorbox{second}{0.141} & \colorbox{second}{0.265} & \colorbox{second}{0.141} & \colorbox{second}{0.868} & \colorbox{best}{0.970} & 
         \colorbox{best}{0.015} & \colorbox{best}{0.006} &\colorbox{second}{0.026} & \colorbox{best}{0.010} & \colorbox{best}{0.694} & \colorbox{best}{0.797} & \colorbox{best}{0.012} & \colorbox{best}{0.006} & \colorbox{second}{0.016} & \colorbox{best}{0.005} & \colorbox{best}{0.900} & \colorbox{best}{0.988} \\
        \bottomrule[0.17em] 
    \end{tabular}
    }
       \caption{
        \textbf{Point Map Estimation.} 
        Keyframes are selected every 5 images for DTU and ETH3D, 40 images for 7-Scenes and 100 images for NRGBD.  We present the accuracy (Acc.), completion (Comp.) and normal consistency (N.C.) as the evaluation metrics with each cell colored to indicate the \colorbox{best}{best} and the \colorbox{second}{second}.
    }
    \label{tab:mv_recon_1}
\end{table*}

\subsection{Monocular Depth Estimation}
For monocular depth estimation, we benchmark on Sintel~\cite{sintel}, Bonn~\cite{bonn}, and NYUv2~\cite{nyuv2} and adopt Absolute Relative Error (Abs Rel $\downarrow$) and threshold accuracy ($\delta<1.25$ $\uparrow$) as evaluation metrics~\cite{monst3r,pi3,moge}.
As reported in Tab.~\ref{tab:monodepth}, our method achieves high-quality results among multi-view approaches and it demonstrates comparable performance with monocular depth estimation model~\cite{moge,mogev2}.

\begin{table}[t]
    \centering
    \tablestyle{1pt}{1.05}
    \resizebox{1.0\columnwidth}!{
    \begin{tabular}{lcccccc}
        \toprule[0.17em]
        \multirow{3}{*}{\textbf{Method}} &
        \multicolumn{2}{c}{\textbf{Sintel~\cite{sintel}}} &
        \multicolumn{2}{c}{\textbf{Bonn~\cite{bonn}}} &
        \multicolumn{2}{c}{\textbf{NYU-v2~\cite{nyuv2}}} \\
        \cmidrule(r){2-3} \cmidrule(r){4-5} \cmidrule(r){6-7} 
        &
        Abs Rel $\downarrow$ & $\delta<1.25\uparrow$ &
        Abs Rel $\downarrow$ & $\delta<1.25\uparrow$ &
        Abs Rel $\downarrow$ & $\delta<1.25\uparrow$ \\
        \midrule
        DUSt3R~\cite{dust3r} & 0.488 & 0.532 & 0.139 & 0.832 &  0.081 & 0.909 \\
        MASt3R~\cite{mast3r} & 0.413 & 0.569 & 0.123 & 0.833 &  0.110 & 0.865 \\
        MonST3R~\cite{monst3r} & 0.402 & 0.525 & 0.069 & 0.954 & 0.094 & 0.887 \\
        Fast3R~\cite{fast3r} & 0.544 & 0.509 & 0.169 & 0.796 & 0.093 & 0.898 \\
        CUT3R~\cite{cut3r} & 0.418 & 0.520 & 0.058 & 0.967  & 0.081 & 0.914 \\
        FLARE~\cite{flare} & 0.606 & 0.402 & 0.130 & 0.836  & 0.089 & 0.898 \\
        VGGT~\cite{vggt} & 0.335 & 0.599 & 0.053 &0.970  & 0.056 & 0.951 \\
        MoGe~\cite{moge} & \colorbox{best}{0.273} &\colorbox{best}{0.695} & 0.050 & 0.976 &  0.055 & 0.952 \\
        Pi3~\cite{pi3} & 0.277
        & 0.614 & \colorbox{best}{0.044} & \colorbox{second}{0.976} & \colorbox{second}{0.054} & \colorbox{second}{0.956} \\
        Ours & \colorbox{second}{0.274} & \colorbox{second}{0.620} & \colorbox{second}{0.050} & \colorbox{best}{0.977} & \colorbox{best}{0.051} & \colorbox{best}{0.957}  \\
        \bottomrule[0.17em]
    \end{tabular}
    }
    \caption{
        \textbf{Monocular Depth Estimation on Sintel, Bonn, and NYU-v2.} We present the absolute relative error (Abs Rel) and threshold accuracy ($\delta<1.25$) as the evaluation metrics with each cell colored to indicate the \colorbox{best}{best} and the \colorbox{second}{second}.
    }
    \label{tab:monodepth}
\end{table}

\subsection{Camera Pose Estimation}
For camera pose estimation, we utilize both angular accuracy and distance error for evaluation.

\noindent\textbf{Angular Accuracy Metrics.} We evaluate predicted poses on the scene-level RealEstate10K~\cite{re10k} and object-centric Co3Dv2~\cite{co3d} datasets following VGGT~\cite{vggt}. For each sequence, we randomly sample 10 images and form all possible pairs. We then compute angular errors of relative rotations and translations, reporting Relative Rotation Accuracy (RRA $\uparrow$) and Relative Translation Accuracy (RTA $\uparrow$). 
We also report a unified metric as the Area Under the Curve (AUC $\uparrow$) of the min(RRA, RTA)–threshold curve. Results in Tab.~\ref{tab:relpose} demonstrate that our method sets a new state of the art on RealEstate10K in the zero-shot setting, while remaining competitive with the best-performing methods on the in-domain Co3Dv2 benchmark.

\noindent\textbf{Distance Error Metrics.} We further assess performance using Absolute Trajectory Error (ATE $\downarrow$), Relative Pose Error for translation (RPE trans $\downarrow$), and rotation (RPE rot $\downarrow$), following CUT3R~\cite{cut3r} using the benchmarks TUM-Dynamics~\cite{tum}..
Predicted trajectories are first aligned with ground truth via a Sim(3) transformation before error computation. As reported in Tab.~\ref{tab:relpose}, our method delivers more accurate results than previous approaches across both synthetic and real-world scenarios.

\begin{table}[t]
    \centering
    \tablestyle{2pt}{1.05}
    \resizebox{1.0\columnwidth}!{
    \begin{tabular}{lccccccccc}
        \toprule
        \multicolumn{1}{l}{\multirow{3}{*}{\textbf{Method}}} &
        \multicolumn{3}{c}{\textbf{RealEstate10K~\cite{re10k}}} &
        \multicolumn{3}{c}{\textbf{Co3Dv2~\cite{co3d}}} &
        \multicolumn{3}{c}{\textbf{TUM-dynamics~\cite{tum}}} \\
        \cmidrule(r){2-4} \cmidrule(r){5-7} \cmidrule(r){8-10}
        \multicolumn{1}{c}{} &
        RRA@30 $\uparrow$ & RTA@30 $\uparrow$ & AUC@30 $\uparrow$ &
        RRA@30 $\uparrow$ & RTA@30 $\uparrow$ & AUC@30 $\uparrow$ &
         ATE $\downarrow$ & RPE trans $\downarrow$ & RPE rot $\downarrow$  \\
        \midrule
        Fast3R~\cite{fast3r} & 99.05 & 81.86 & 61.68 & 97.49 & 91.11 & 73.43 & 0.090 & 0.101 & 1.425\\
        CUT3R~\cite{cut3r} & 99.82 & 95.10 & 81.47 & 96.19 & 92.69 & 75.82 & 0.047 & 0.015 & 0.451\\
        FLARE~\cite{flare} & 99.69 & 95.23 & 80.01 & 96.38 & 93.76 & 73.99 & 0.026 & 0.013 & 0.475\\
        VGGT~\cite{vggt} & 99.97 & 93.13 & 77.62 & 98.96 & 97.13 & \colorbox{best}{88.59} & \colorbox{second}{0.012} & 0.010 & \colorbox{second}{0.311}\\
        Pi3~\cite{pi3} & \colorbox{best}{99.99} & \colorbox{best}{95.62} & \colorbox{second}{85.90} & \colorbox{second}{99.05} & \colorbox{second}{97.33} & 88.41 & 0.014 & \colorbox{second}{0.009} & 0.312\\
        Ours & \colorbox{second}{99.98} & \colorbox{second}{95.39} & \colorbox{best}{86.13} & \colorbox{best}{99.08} & \colorbox{best}{97.36} & \colorbox{second}{88.42} &  \colorbox{best}{0.011} & \colorbox{best}{0.009} &  \colorbox{best}{0.311} \\
        \bottomrule
    \end{tabular}
    }
    \caption{\textbf{Camera Pose Estimation on RealEstate10K, Co3Dv2 and TUM-dynamics.}
    We present the metrics that measure the ratio of angular accuracy of rotation/translation under an error of 30 degrees for RealEstate10K and Co3Dv2. We also present the distance error of rotation/translation for TUM-dynamics with each cell colored to indicate the \colorbox{best}{best} and the \colorbox{second}{second}.
    }
    \label{tab:relpose}
\end{table}

\begin{figure*}
    \centering
    \includegraphics[width=\linewidth]{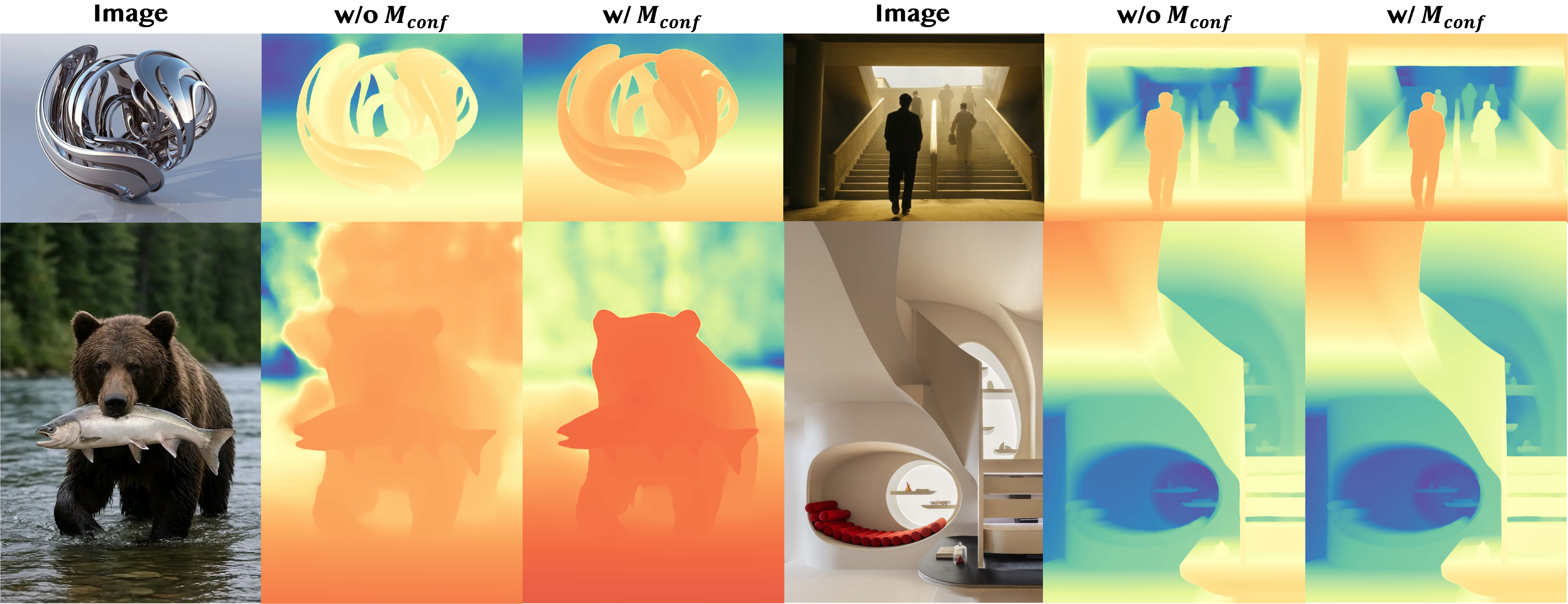}
    \caption{Ablation for confidence-based depth refinement. We demonstrate the effectiveness of the confidence-based depth refinement for more accurate depth estimation.}
    \label{fig:depth_ab}
\end{figure*}

\begin{figure*}
    \centering
    \includegraphics[width=\linewidth]{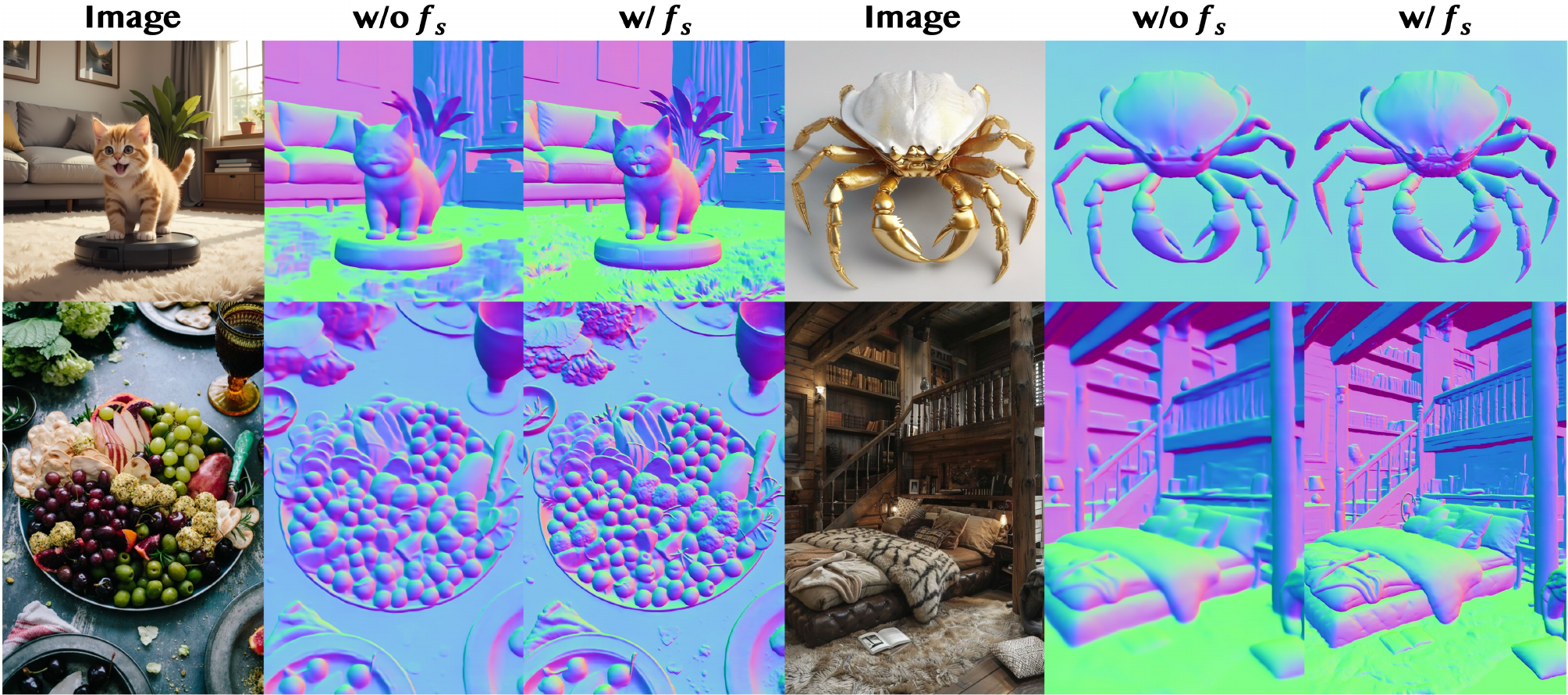}
    \caption{Ablation for dense semantic feature fusion. We demonstrate the effectiveness of the dense semantic feature fusion for sharper and more accurate normal estimation.}
    \label{fig:normal_ab}
\end{figure*}

\subsection{Normal Estimation}
For normal estimation, we evaluate our model on several benchmark datasets spanning both indoor and outdoor scenes. We compare against Magrigold~\cite{ke2023repurposing}, Lotus~\cite{lotus}, GeoWizard~\cite{geowizard}, and StableNormal~\cite{stablenormal}. 
As shown in Tab.\ref{tab:normal_result}, our model achieves superior performance across multiple benchmarks. 
By alleviating the inherent ambiguity of monocular estimation and incorporating dense semantic features, our method produces more reliable and detailed predictions across diverse scenarios.

\begin{table}
\centering
\tablestyle{2pt}{1.05}{
\resizebox{1.0\columnwidth}!{
  \begin{tabular}{lccccccccc}
    \toprule[0.17em]
    \multicolumn{1}{l}{\multirow{3}{*}{\textbf{Method}}} 
    & \multicolumn{3}{c}{\textbf{NYUv2}~\cite{nyuv2}} & \multicolumn{3}{c}{\textbf{ScanNet}~\cite{yeshwanthliu2023scannetpp}} & \multicolumn{3}{c}{\textbf{IBims-1}~\cite{ibims}}   \\
        \cmidrule(r){2-4} \cmidrule(r){5-7} \cmidrule(r){8-10}
        \multicolumn{1}{c}{} &
     Mean $\downarrow$ & Med $\downarrow$ & $\delta_{11.25^\circ}$ $\uparrow$ & Mean $\downarrow$ & Med $\downarrow$ & $\delta_{11.25^\circ}$ $\uparrow$ & Mean $\downarrow$ & Med $\downarrow$ & $\delta_{11.25^\circ}$ $\uparrow$   \\
    \midrule
    Marigold~\cite{ke2023repurposing} & 20.8 & 11.1 & 50.4 & 21.2 & 12.2 & 45.6 & 18.4 & 8.4 & 64.7  \\
    Lotus~\cite{lotus} & \colorbox{second}{17.5} & \colorbox{second}{8.6} & \colorbox{second}{58.7} & 18.1 & \colorbox{second}{8.8} & \colorbox{second}{58.2} & 19.2 & \colorbox{second}{5.6} & 66.2  \\
    GeoWizard~\cite{geowizard} & 20.4 & 11.9 & 47.0 & 21.4 & 13.9 & 37.1 & 19.7 & 9.7 & 58.4 \\
    StableNormal~\cite{stablenormal} & 19.7 & 10.5 & 53.0 & \colorbox{second}{18.1} & 10.1 & 56.0 & \colorbox{second}{17.2} & 8.1 & \colorbox{second}{66.7}   \\
    Ours & \colorbox{best}{15.1} & \colorbox{best}{7.3} &  \colorbox{best}{63.5} & \colorbox{best}{16.1} & \colorbox{best}{7.2} &  \colorbox{best}{64.4} &\colorbox{best}{15.0} & \colorbox{best}{4.2} &  \colorbox{best}{72.6}  \\
    \bottomrule[0.17em]
\end{tabular}}}
  \caption{Quantitative comparison of normal prediction. We report the mean and median angular errors with each cell colored to indicate the \colorbox{best}{best} and the \colorbox{second}{second}.}
  \label{tab:normal_result}
\end{table}

\subsection{Ablation Study}
\noindent\textbf{Mixtures-of-Experts.}
To validate the effectiveness of our proposed framework, we perform an ablation study by removing key components and training corresponding model variants. We compare the model variants by evaluating their performance on pointmap estimation (DTU~\cite{dtu}), monocular depth prediction (NYUv2~\cite{nyuv2}), and camera pose (RealEstate10K~\cite{re10k}).
As shown in Tab.~\ref{tab:moe}, the results confirm that both the MoE and the tailored loss (including the confidence-based depth refinement) play a crucial role, contributing significantly to the overall performance of our model. 
It is noteworthy that all model variants, including the baseline, are trained for the same number of steps.

\begin{table}
\centering
\tablestyle{2pt}{1.05}{
    \resizebox{1\linewidth}{!}{
    \begin{tabular}{c|ccc|cc|ccc}
    \toprule
    \multirow{2}{*}{\textbf{Model}} 
    & \multicolumn{3}{c|}{\textbf{DTU}~\cite{dtu}} 
    & \multicolumn{2}{c|}{\textbf{NYUv2}~\cite{nyuv2}} 
    & \multicolumn{3}{c}{\textbf{RealEstate10K}~\cite{re10k}} \\
    \cmidrule(r){2-4} \cmidrule(r){5-6} \cmidrule(r){7-9} 
    & Acc. $\downarrow$ & Comp. $\downarrow$ & N.C. $\uparrow$ 
    & Abs Rel $\downarrow$ & $\delta < 1.25 \uparrow$ 
    & RRA@30 $\uparrow$ & RTA@30 $\uparrow$ & AUC@30 $\uparrow$ \\
    \midrule
    w/o $\mathcal{L}$, w/o MoE & 1.338 & 1.896 & 0.676 & 0.056 & 0.951 & 99.97 & 93.13 & 77.62 \\
    w/o MoE  & 1.297 & 1.625 & 0.682 & 0.054 & 0.953 & 99.94 & 94.27 & 85.14  \\
    Ours                        & \textbf{1.011} & \textbf{1.491} & \textbf{0.695} & \textbf{0.051} & \textbf{0.957} & \textbf{99.98} & \textbf{95.39} & \textbf{86.13} \\
    \bottomrule
    \end{tabular}}
    }
\caption{Ablation study on the key components of our model. The results illustrate how performance metrics improve progressively as each component is incorporated into the baseline (w/o $\mathcal{L}$, w/o MoE). Note that we report the \textit{Mean} values of Acc., Comp., and N.C. for the ETH3D dataset (pointmap estimation).}
  \label{tab:moe}
\end{table}

\noindent\textbf{Confidence-based Depth Refinement.}
For depth prediction, we design a confidence-based depth refinement that utilizes MoGev2 to filter depth supervision.
By restricting supervision to high-confidence regions, our model avoids overfitting to corrupted data and achieves more accurate depth estimation, which can be shown in Tab.~\ref{tab:moe} and Fig.~\ref{fig:depth_ab}.

\noindent\textbf{Dense Semantic Feature Fusion.}
MoRE proposes to fuse globally aligned 3D backbone features with dense semantic features for normal predictions. This feature fusion allows the model to capture both local geometric details and global contextual cues, leading to sharper and more reliable predictions.
As shown in Fig.~\ref{fig:normal_ab}, it further strengthens the representation of fine-grained structures, which is essential for high-fidelity 3D reconstruction.
\section{Conclusion}
\label{sec:con}
We present MoRE, a large-scale 3D visual foundation model that integrates a Mixture-of-Experts (MoE) framework to enable task-specialized feature learning for multiple 3D geometric prediction tasks.
Unlike previous methods that rely on a single shared representation for all scenarios, MoRE dynamically routes features to task-specific experts, allowing each head to learn specialized representations and improve prediction accuracy. 
To address the noise and inconsistency of real-world data, our model employs a confidence-based depth refinement module, which effectively enhances the reliability of depth estimation.
We further introduce a dense semantic fusion that fuses local geometry features with 3D backbone features for more detailed normal estimation.
The model is trained with tailored multi-task losses and an adaptive loss mechanism to stabilize training across generalized inputs and diverse datasets. 
Extensive experiments demonstrate that MoRE achieves state-of-the-art performance across multiple 3D prediction benchmarks, providing a versatile and scalable backbone for downstream 3D vision applications.
{
    \small
    \bibliographystyle{ieeenat_fullname}
    \bibliography{main}
}
\clearpage \appendix

\section{Appendix}
\label{sec:appendix}
\subsection{Additional Visualization}
\begin{figure*}
    \centering
    \includegraphics[width=\linewidth]{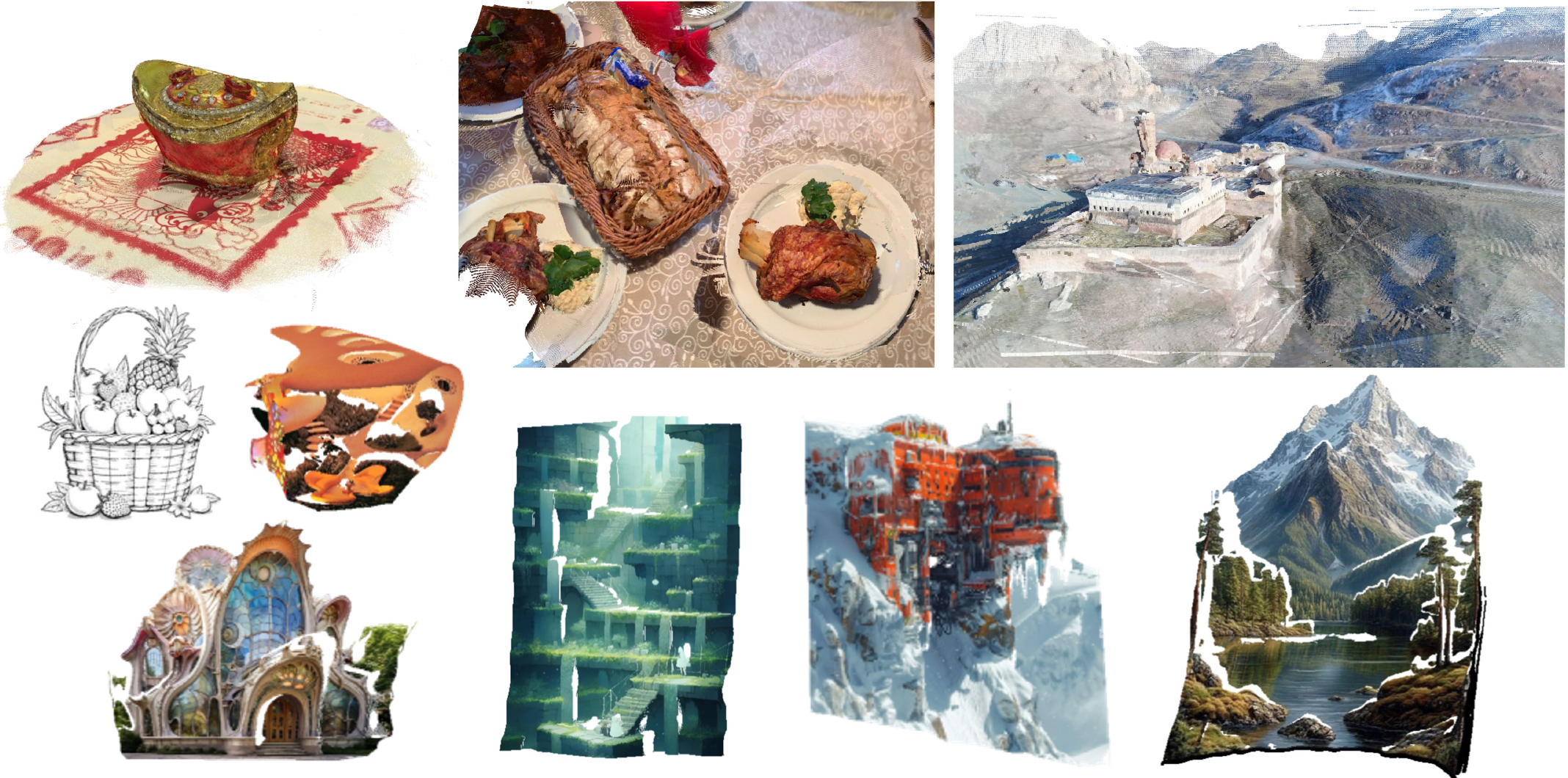}
    \caption{Additional pointmap visualization. MoRE achieves highly accurate and realistic reconstructions for both real-world (first-row) and synthetic inputs (second-row).}
    \label{fig:pointmap_vis}
\end{figure*}

We present additional visualizations of the reconstructed pointmaps in Fig.~\ref{fig:pointmap_vis}, including results from both real-world (first-row) and synthetic inputs (second-row). MoRE produces highly accurate and realistic reconstructions in a single forward pass, demonstrating its potential to further benefit downstream 3D applications.

\subsection{Additional Comparison}
\begin{figure*}
    \centering
    \includegraphics[width=\linewidth]{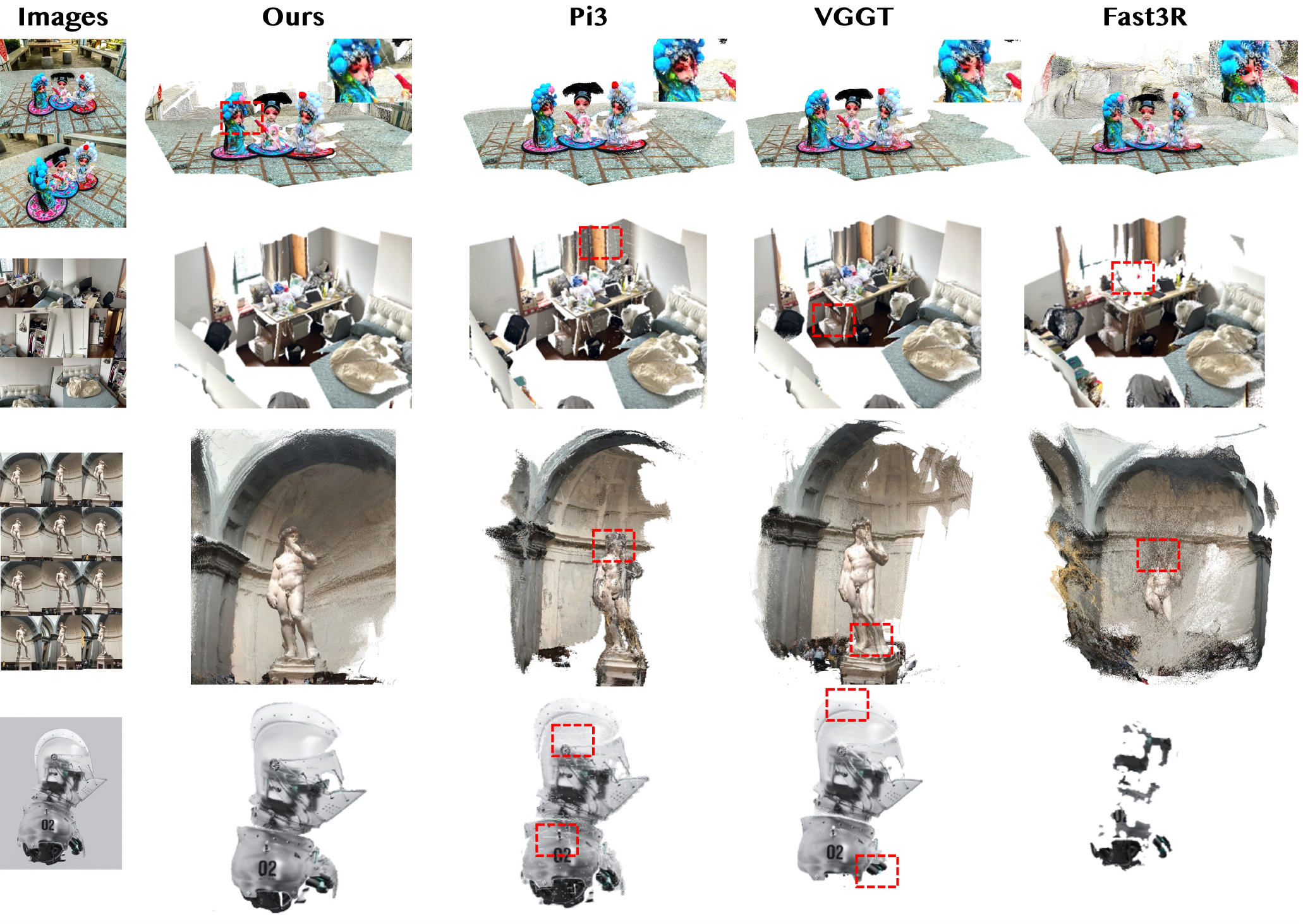}
    \caption{Additional pointmap comparison. Our method can produce more accurate pointmaps and capture more realistic structural details.}
    \label{fig:pointmap_sup}
\end{figure*}

\noindent\textbf{Pointmap Comparison.}
We provide additional pointmap comparisons in Fig.~\ref{fig:pointmap_sup}, where our method reconstructs more accurate pointmaps and captures more realistic structural details. 

\begin{figure*}
    \centering
    \includegraphics[width=\linewidth]{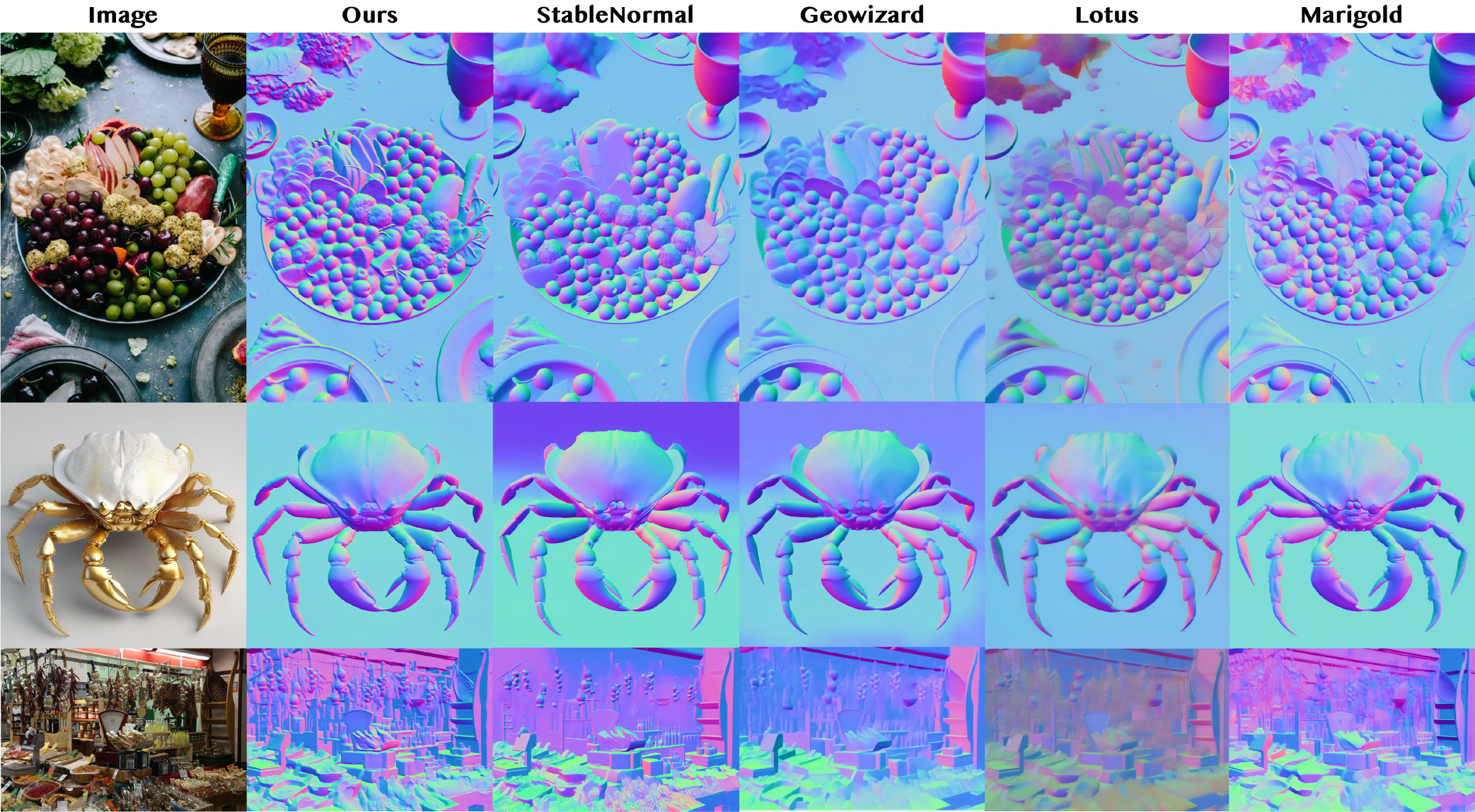}
    \caption{Normal comparison. Our method produces sharper and more accurate surface normals.}
    \label{fig:normal_comp}
\end{figure*}

\noindent\textbf{Normal Comparison.} 
We also present normal prediction comparisons with previous methods in Fig.~\ref{fig:normal_comp}, showing that our approach produces sharper and more accurate surface normals.

\noindent\textbf{Video Depth Comparison.} 
We also provide video depth estimation comparison in Tab.~\ref{tab:videodepth}. It can be seen that MoRE can also achieve accurate depth estimation for dynamic scenarios.

\begin{figure}
    \centering
    \includegraphics[width=\linewidth]{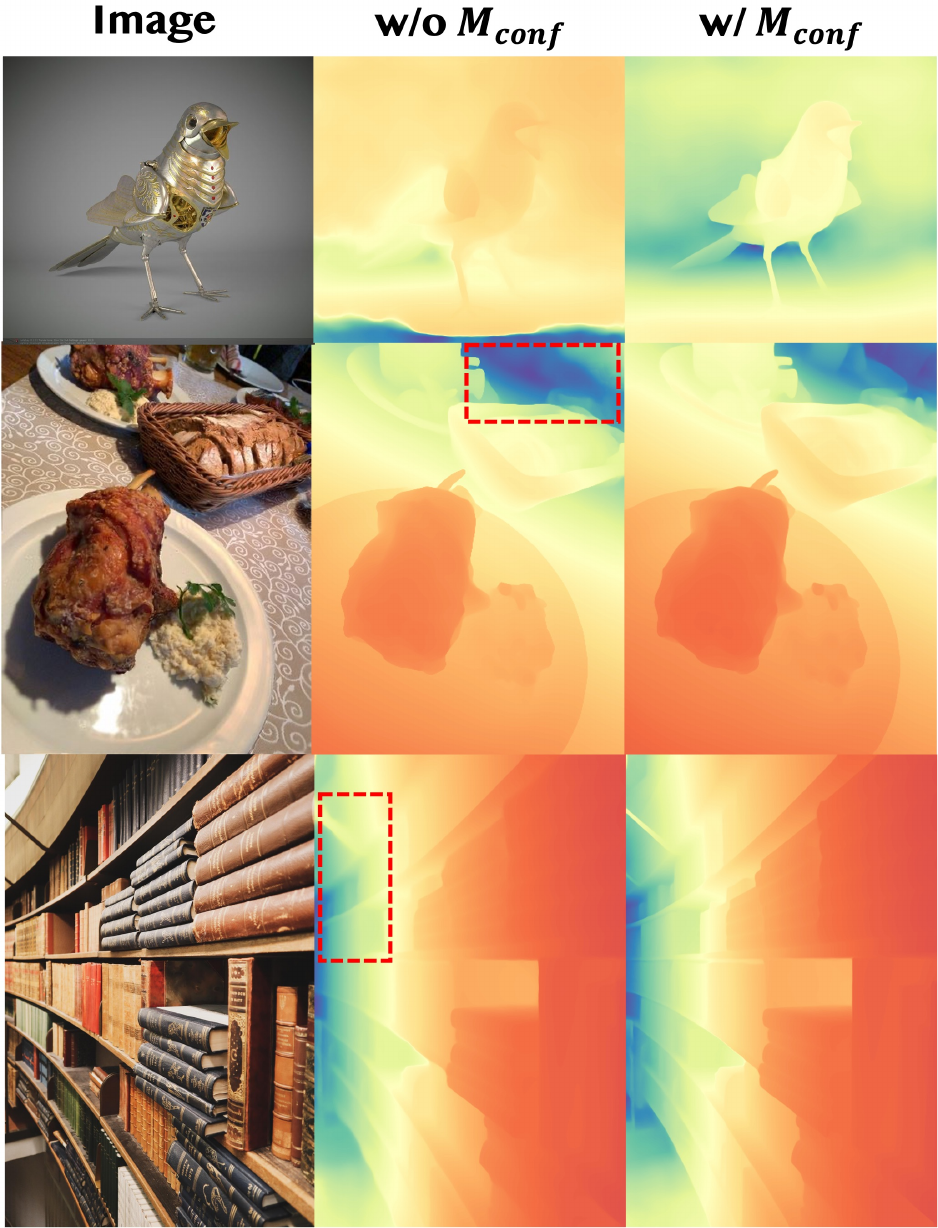}
    \caption{Additional ablation for confidence-based depth refinement. Please zoom in for better details.}
    \label{fig:depth_ab_sup}
\end{figure}

\begin{figure*}
    \centering
    \includegraphics[width=\linewidth]{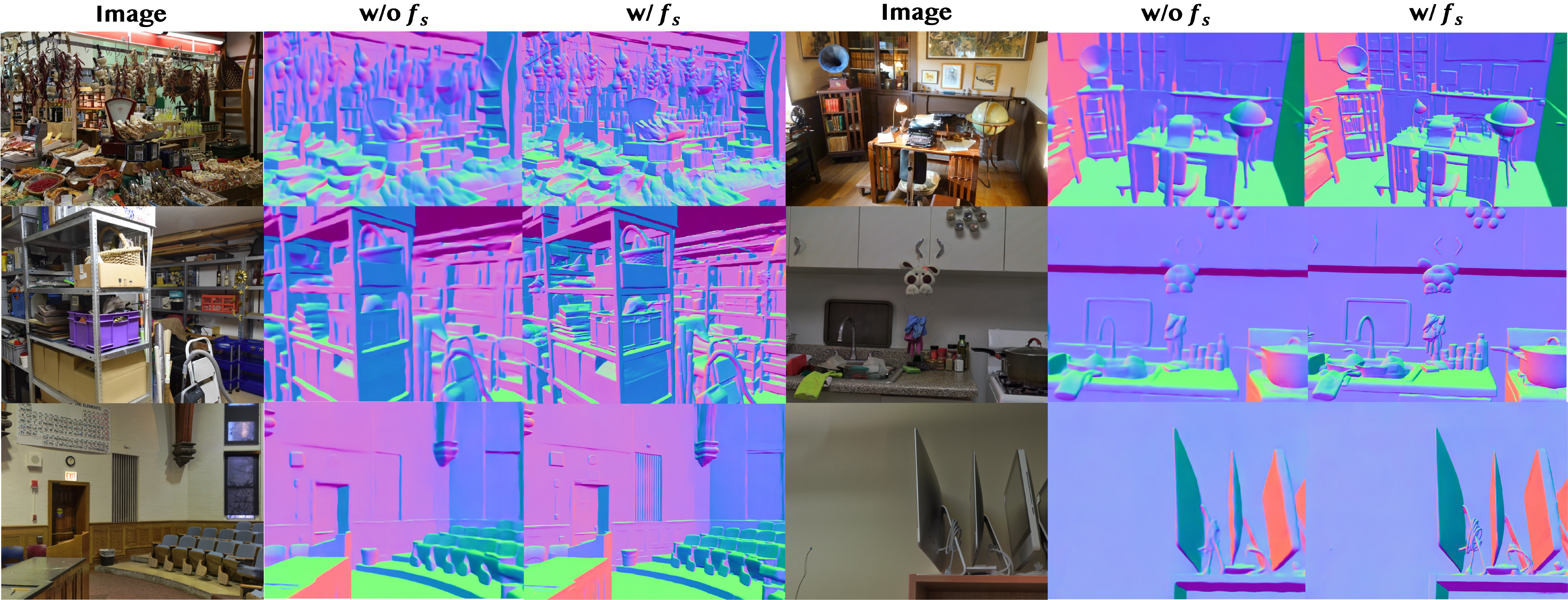}
    \caption{Additional ablation for dense semantic feature fusion. Please zoom in for better details.}
    \label{fig:normal_ab_sup}
\end{figure*}

\noindent\textbf{Ablation Study.} 
Furthermore, Fig.~\ref{fig:depth_ab_sup} present additional ablation results on confidence-based depth refinement, demonstrating that the proposed design effectively enhances depth accuracy.
In addition, Fig.~\ref{fig:normal_ab_sup} and Tab.~\ref{tab:ablation_normal} present ablation results on dense semantic fusion for normal prediction, demonstrating that the proposed fusion effectively contributes to sharper and more precise normal estimations.

\begin{table}[h]
    \centering
      \tablestyle{2pt}{1.05}
    \resizebox{\columnwidth}!{
\begin{tabular}{c|cc|cc|cc}
\hline
& \multicolumn{2}{c|}{\textbf{NYUv2}~\cite{nyuv2}}                            & \multicolumn{2}{c|}{\textbf{ScanNet}~\cite{yeshwanthliu2023scannetpp}}                          & \multicolumn{2}{c}{\textbf{IBims}~\cite{ibims}}                                                    \\
& Mean $\downarrow$ & $\delta_{11.25 \degree}\uparrow $ & Mean $\downarrow$ & $\delta_{11.25 \degree}\uparrow $ & Mean $\downarrow$ & $\delta_{11.25 \degree}\uparrow $   \\ \hline
w/o $f_s$ & 15.6 & 63.0  & 16.5 & 63.8 & 15.5 & 71.2  \\
Ours    & \textbf{15.1} & \textbf{63.5} & \textbf{16.1} & \textbf{64.4} & \textbf{15.0} & \textbf{72.6} \\\hline
\end{tabular}}
    \caption{Normal quantitative metrics for ablation. We demonstrate that the dense semantic feature fusion ($f_s$) contributes to more accurate normal predictions.}
  \label{tab:ablation_normal}
\end{table}

\begin{table}
    \centering
    \tablestyle{2pt}{1.05}
    \resizebox{\columnwidth}!{
    \begin{tabular}{lcccccc}
        \toprule
        {\multirow{3}{*}{\textbf{Method}}} &
        \multicolumn{2}{c}{\textbf{Sintel}~\cite{sintel}} &
        \multicolumn{2}{c}{\textbf{Bonn}~\cite{bonn}} &
        \multicolumn{2}{c}{\textbf{KITTI}~\cite{kitti}} \\
        \cmidrule(r){2-3} \cmidrule(r){4-5} \cmidrule(r){6-7}
        & 
        Abs Rel $\downarrow$ & $\delta<1.25$ $\uparrow$ &
        Abs Rel $\downarrow$ & $\delta<1.25$ $\uparrow$ &
        Abs Rel $\downarrow$ & $\delta<1.25$ $\uparrow$ \\
        \midrule
        DUSt3R~\cite{dust3r} & 0.570 & 0.493 & 0.152 & 0.835 & 0.135 & 0.818 \\
        MASt3R~\cite{mast3r}  & 0.480 & 0.517 & 0.189 & 0.771 & 0.115 & 0.849 \\
        MonST3R~\cite{monst3r}  & 0.402 & 0.526 & 0.070 & 0.958 & 0.098 & 0.883  \\
        Fast3R~\cite{fast3r}  & 0.518 & 0.486 & 0.196 & 0.768 & 0.139 & 0.808  \\
        MVDUSt3R~\cite{mvdust3r}  & 0.619 & 0.332 & 0.482 & 0.357 & 0.401 & 0.355  \\
        CUT3R~\cite{cut3r}  & 0.534 & 0.558 & 0.075 & 0.943 & 0.111 & 0.883 \\
        FLARE~\cite{flare}  & 0.791 & 0.358 & 0.142 & 0.797 & 0.357 & 0.579 \\
        VGGT~\cite{vggt}  & 0.230 & 0.678 & 0.052 & 0.969 & 0.052 & 0.968 \\
        Pi3~\cite{pi3}  & \colorbox{second}{0.210} & \colorbox{second}{0.726} & \colorbox{second}{0.043} & \colorbox{second}{0.975} & \colorbox{best}{0.037} & \colorbox{best}{0.985} \\
         Ours  & \colorbox{best}{0.205} & \colorbox{best}{0.728} & \colorbox{best}{0.042} & \colorbox{best}{0.975} & \colorbox{second}{0.047} & \colorbox{second}{0.978} \\
        \bottomrule
    \end{tabular}
    }
      \caption{
        \textbf{Video Depth Estimation on Sintel~\cite{sintel}, Bonn~\cite{bonn} and KITTI~\cite{kitti}}.We present the absolute relative error (Abs Rel) and threshold accuracy ($\delta<1.25$) as the evaluation metrics with each cell colored to indicate the \colorbox{best}{best} and the \colorbox{second}{second}.
    }
    \vspace{-1em}
    \label{tab:videodepth}
\end{table}

\end{document}